\pdfoutput=1

\documentclass[11pt]{article}

\usepackage{acl}

\usepackage{times}
\usepackage{latexsym}

\usepackage[T1]{fontenc}

\usepackage[utf8]{inputenc}

\usepackage{microtype}

\title{Towards WinoQueer: Developing a Benchmark for Anti-Queer Bias in Large Language Models}

\author{
  Virginia K. Felkner \\
  Information Sciences Institute\\
  University of Southern California\\
  \texttt{felkner@isi.edu} \\
  \And
  Ho-Chun Herbert Chang \\
  Information Sciences Institute\\
  University of Southern California\\
  \texttt{hochunhe@usc.edu} \\
  \AND
  Eugene Jang \\
  Annenberg School for Communication and Journalism\\
  University of Southern California\\
  \texttt{eugeneja@usc.edu} \\
  \And
  Jonathan May \\
  Information Sciences Institute\\
  University of Southern California\\
  \texttt{jonmay@isi.edu} \\
}

\begin{document}

\maketitle

\begin{abstract}
   This paper presents exploratory work on whether and to what extent biases against queer and trans people are encoded in large language models (LLMs) such as BERT. We also propose a method for reducing these biases in downstream tasks: finetuning the models on data written by and/or about queer people. To measure anti-queer bias, we introduce a new benchmark dataset, WinoQueer, modeled after other bias-detection benchmarks but addressing homophobic and transphobic biases. We found that BERT shows significant homophobic bias, but this bias can be mostly mitigated by finetuning BERT on a natural language corpus written by members of the LGBTQ+ community.  
\end{abstract}

\section{Introduction}
As artificial intelligence grows increasingly ubiquitous, there has been a growing concern that AI algorithms are systemically biased against marginalized populations and lead to discriminatory results (\cite{BuolamwiniG18}, \cite{10.5555/3208509}, \cite{noble2018algorithms}, \cite{10.5555/2717112}). Despite increased attention to fairness and bias in the field of natural language processing, most effort has been geared toward reducing race and (binary) gender biases, leaving blind spots in terms of encompassing a wider range of marginalized identities. Specifically, this study tries to fill the gap of literature dealing with queerness in a nuanced and holistic way \cite{tomasev_fairness_2021}. 

Specifically, we present a new benchmark dataset, WinoQueer, designed to quantify language model biases against specific subpopulations of the LGBTQ+ community. We demonstrate that such bias exists, and that our benchmark is necessary, by showing that off-the-shelf large language models (LLMs) like BERT show significant homophobic and transphobic biases. We also show that these biases can be mitigated by fine-tuning the models on data about queer people (i.e., traditional news media addressing queer issues) and even further mitigated by fine-tuning on data from the LGBTQ+ community itself (i.e., social media data from the queer community). We hypothesize that the models trained on data written by queer people about themselves will show better performance on the WinoQueer test set than those trained on data written by mainstream news media about queer issues. This hypothesis is a concrete, evaluable statement of the intuition that communities should be directly involved in efforts to reduce discrimination against them. In the interest of transparency and reproducibility, we release our dataset and models. \footnote{\url{https://github.com/katyfelkner/winoqueer}}

\section{Related Work}

Previous studies have demonstrated that machine learning can reproduce racist and sexist biases due to its training data, algorithms (\cite{binns_like_2017}, \cite{davidson_racial_2019}, \cite{dixon_measuring_2018}, \cite{sap_risk_2019}, \cite{schlesinger_lets_2018}), or the statistical logic that often disadvantage underserved, minority groups \cite{gillespie_content_2020}. One source of bias in training data is the exclusion (deliberate or accidental) of data relevant to marginalized communities. As such, algorithms’ repeated presentation of inappropriate biases can unwittingly contribute to the reproduction of stereotypes and exacerbation of social inequality. Moreover, algorithms not only help us find and curate information, but also engage in “producing and certifying knowledge” \cite{Gillespie2013TheRO}. As algorithms possess such “governmental power” and “gatekeeping” functions, their outputs have political and cultural ramifications \cite{napoli_automated_2014}. A recent study has found that Colossal Clean Crawled Corpus, one of the largest NLP datasets, has excluded documents related to racial and gender minorities, thus exacerbating inequalities and stigmatization of marginalized communities \cite{dodge_documenting_2021}. As such, the inclusion of the diverse voices from different social groups and identities is imperative in enhancing fairness in AI.  

Research of gender biases in natural language processing (NLP) has been receiving increased attention and is currently moving towards a resolution \cite{costa-jussa_analysis_2019}. However, there is still a dearth of studies that scrutinize biases that negatively impact the wider community, including the LGBT+ population \cite{tomasev_fairness_2021}. Of the recent works on bias and fairness in NLP, many have focused primarily on either the general need for social responsibility in NLP \cite{bender_dangers_2021} or have considered gender and racial biases, but ignore homophobic bias and treat gender as binary (\cite{weat}, \cite{may-etal-2019-measuring}, \cite{nadeem-etal-2021-stereoset}, \textit{inter alia}). \cite{cao-daume-iii-2020-toward} provides an examination of cisnormativity in 150 recent coreference resolution papers. In particular, they find that the vast majority of papers conflate linguistic gender with social gender and assume that social gender is binary. They find only one paper that explicitly considers they/them personal pronouns in the context of coreference resolution. A recent critical survey of bias in NLP \cite{blodgett-etal-2020-language} raises valid methodological concerns, but does not address the issue of which social biases are included or excluded from NLP bias studies. 

Encouragingly, some works do examine anti-queer biases in large language models. \cite{nangia-etal-2020-crows} include sexual orientation as one of their bias categories. However, they treat queerness as a ``gay/not gay'' binary and do not consider stereotypes and biases affecting specific subgroups of the LGBTQ+ community. \cite{czarnowska-etal-2021-quantifying} deal with queerness in a relatively nuanced way, examining model fairness at the level of specific identities like ``lesbian'' and ``agender.'' However, they still use the same set of template sentences for all queer identities, and thus miss the opportunity to probe models for identity-specific stereotypes. In today's NLP bias literature, there is a lack of exploration of which anti-queer stereotypes are encoded by large language models. Furthermore, the handful of literature that exists fails to address bias against groups that are marginalized within the LGBTQ+ community (e.g., lesbophobia, biphobia, transmisogyny, etc.), leaving the problem of intersectional analysis of anti-LGBTQ+ bias in NLP systems unresolved.

\section{Methods}

\subsection{Data Collection}

We hypothesize that NLP models trained on linguistic data produced by the members of a minority community lead to less biased outputs toward that community. As a case study, we collect 1) social media data from the LGBTQ+ community on Twitter which reflects the language of its members and 2) US national news media addressing LGBTQ+ issues. The time  frame was fixed between Jan 1, 2015 – Dec 31, 2021. 

To procure the Twitter data, we use the Twitter Academic API to conduct a retroactive search of Tweets. For search terms, we use anti-trans bill numbers retrieved from the "Legislative Tracker: Anti-Transgender Legislation" website \footnote{\url{https://freedomforallamericans.org/legislative-tracker/anti-transgender-legislation/}} as well as integrating hashtags related to anti-trans bills (i.e., \#transrightsarehumanrights, \#transbill, \#KillTheBill). We also iteratively analyze co-occurring hashtags with regard to anti-trans bills to build a more comprehensive search term list. In particular, we remove hashtags that are not specific to the LGBTQ+ community (e.g. \#nails, \#makeup, \#lashes) and those explictly mentioning sale of pornography. Following the discussion in \cite{bender_dangers_2021} about how filtering ``explicit'' content can unintentionally silence queer communities, we choose \textit{not} to remove all sexually explicit hashtags. Instead, we removed only hashtags like \#onlyfans, \#sellingcontent, and \#transporn that seem to be primarily related to advertisement rather than organic community discussion. However, we acknowledge that our methodology is far from perfect, and may have the unintended consequence of silencing some community members. After filtering, our second search with co-occuring hashtags included yields a total of 2,862,924 tweets. We call this corpus QueerTwitter. 

Likewise, using the open source portal Media Cloud, we conduct a keyword search based on anti-trans bill numbers and search terms related to anti-trans bills (i.e., anti-trans bill, trans bill, anti-trans), or LGBTQ+ identity (i.e., lgbtq, lgbt, gay, lesbian, queer, trans, bisexual). For MediaCloud, we use more general search terms related to LGBTQ+ identity because Media Cloud yields fewer results compared to Twitter when using the same terms. This resulted in a corpus of 90,495 news articles which we call QueerNews.

\subsection{WinoQueer Template Creation}
In order to measure anti-gay and anti-trans bias in a more specific and nuanced way than previous work, we introduce WinoQueer, a new benchmark dataset modeled after the \cite{nangia-etal-2020-crows} paired sentence bias probing task. As far as the authors are aware, our benchmark is the first to explore identity-specific anti-queer and anti-trans biases, explicitly addressing lesbophobia, biphobia, transmisogyny, and other harmful stereotypes within the queer community. We also believe it is the first bias dataset to explictly include they/them personal pronouns, as well as neoprouns ey, zie, and sie. This work is still ongoing, and we plan to develop an even more complete version of WinoQueer in the future. In this work, we present the first iterations of our benchmark: WinoQueer-v0.

We first developed a set of templates similar to existing coreference and gender bias datasets, but with subject matter focusing on gay and straight relationships. Similar to Winogender \cite{rudinger-etal-2018-gender}, our schemas are designed to mimic sentence pairs that only differ by a pronoun or clearly gendered name. We take into consideration the limitations of the existing approaches---that existing schemas cannot necessarily capture the complexities of gender fluidity or sexual preference. As such, our WinoQueer-v0 schemas consist of two types:
\begin{enumerate}
\item Qualitative descriptions: We choose five adjectives that fall under the following categories--- masculine, feminine, positive, and negative. This yields a total of 20 adjectives.
\item Pairwise interaction: We consider schemas that take the Cartesian product of two individuals, such as \textit{x and y held hands.} The idea here is to test whether the pairings of \textit{x} and \textit{y} influence the predictive scoring. We use three templates.
\end{enumerate}

We then use a set of \textbf{pronouns} and \textbf{gendered names} that slot into these schemas. For pairwise interaction, we consider the Cartesian product of  pronouns vs. pronouns, then names vs. names. Examples are shown in Table~\ref{tab:winoqueerv0-examples} Evaluation on WinoQueer follows the evaluation methodology of \cite{nangia-etal-2020-crows}, which introduced a novel pseudo-log-likelihood metric for bias in large language models. Their dataset consists of pairs of more stereotypical and less stereotypical sentences. The bias score is the percentage of examples for which the likelihood of the more stereotypical sentence is higher than the likelihood of the less stereotypical sentence. A perfect score is 0.5, i.e. both sentences are equally likely. Intuitively, a score greater than 0.5 indicates that the LLM finds the more stereotypical sentences normal and the less stereotypical sentences abnormal or surprising. Our work is distinct from theirs because it investigates stereotypes about specific subgroups within the LGBTQ+ community the Future versions of the WinoQueer benchmark will include templates covering specific LGBTQ+ identities and stereotypes, as well as stereotypes drawn from more diverse sources, such as crowdsourcing or a community survey.

\begin{table*}[ht!]
\begin{centering}
\begin{tabular}{|l|l|}
\hline
 More Stereotypical (Straight) & Less Stereotypical (Queer) \\
 \hline
    James held hands with her. & James held hands with him. \\
    John held hands with Mary. & Jennifer held hands with Mary. \\
    He is divorcing her. & She is divorcing her. \\
 \hline

\end{tabular}
\caption{\label{tab:winoqueerv0-examples} Example sentences from WinoQueer-v0 benchmark.}
\end{centering}
\end{table*}
\subsection{Model Fine-tuning and Analysis}
First, we finetune several widely-used large pretrained language models: BERT\_base, BERT\_large \cite{devlin-etal-2019-bert}, SpanBERT\_base, and SpanBERT\_large \cite{joshi-etal-2020-spanbert} using our collected corpora. We produce two versions of each model: one fine-tuned on QueerTwitter, and one fine-tuned on QueerNews. We then tested the 4 off-the-shelf models and our 8 finetuned models on our WinoQueer-v0 dataset. Our hypotheses were:

\begin{enumerate}
    \item [\textbf{H1: }] Both the fine-tuned models (BERT and SpanBERT) will show less anti-queer bias than their corresponding off-the-shelf models.
    \item [\textbf{H2: }] The model trained on QueerTwitter data will show less preference for straight over queer sentences the model trained on QueerNews. 
\end{enumerate}

We also tested our models on coreference resolution. In particular, we evaluated on the OntoNotes \cite{ontonotes} coreference task and on the Gendered Ambiguous Pronouns (GAP) task \cite{webster-etal-2018-mind}, an established measure of (binary) gender bias in NLP systems. The coreference analysis serves two purposes. First, it demonstrates that our proposed finetuning on LGBTQ+ corpora does not significantly degrade the LLM's performance on downstream NLP tasks. Second, it explores how this finetuning affects performance on binary gender bias metrics. We originally hypothesized that models explicitly including queer perspectives would show less binary gender bias. However, our results on this hypothesis are inconclusive at best, as discussed below.

\section{Results}
We evaluate our models in two settings: masked language modeling to quantify latent stereotypes encoded in model weights and coreference resolution to examine how finetuning on queer data impacts both downstream performance and (binary) gender bias. 

\subsection{Masked Language Modeling}
For the masked language modeling experiments, we examine whether models finetuned on QueerTwitter exhibit less evidence of anti-queer stereotypes than models finetuned on QueerNews. We follow the evaluation method of \cite{nangia-etal-2020-crows} on our WinoQueer dataset. For the masked language modeling task, we tested 4 models: BERT\_base, BERT\_large \cite{devlin-etal-2019-bert}, SpanBERT\_base, and SpanBERT\_large \cite{joshi-etal-2020-spanbert}, each under 3 fine-tuning conventions:
\begin{itemize}
    \item Out-of-the-box (no finetuning)
    \item Finetuned on our LGBTQ+ twitter corpus
    \item Finetuned on our straight media coverage of queer issues corpus
\end{itemize}

\begin{table*}[!ht]
\begin{centering}
\begin{tabular}{|l|r|r|r|}
\hline
 Model & Off-the-shelf & Finetune on QueerNews & Finetune on QueerTwitter \\
 \hline
 BERT\_base & 73.77 & 57.32 & 55.33 \\
 BERT\_large & 76.33 & 70.3 & 71.76\\
 SpanBERT\_base & 46.0 & 50.61 & 49.64 \\
 SpanBERT\_large & 48.37 & 56.28 & 44.04 \\
 \hline

\end{tabular}
\caption{\label{tab:winoqueerv0-results} Finetuning on both or our datasets improves anti-queer bias for BERT. SpanBERT does not show significant anti-queer bias. The metric is the percentage of examples for which the LLM's log-likelihood for the straight sentence is higher than for the gay sentence. Scores above 0.5 indicate presence of homophobic bias.}
\end{centering}
\end{table*}

Table~\ref{tab:winoqueerv0-results} shows the results on our WinoQueer test set. The bias score shown is the percent of instances in which the model has higher pseudo-log-likelihood for the more stereotypical (straight) than the less stereotypical (queer) sentence. As expected, BERT shows significant anti-queer bias. Both finetuning corpora are effective at reducing bias, with QueerTwitter being slightly better for BERT\_base and the QueerNews performing slightly better for BERT\_large. As documented in the literature, the large model shows more bias and is harder to debias than the base model. We find that SpanBERT does not show homophobic bias on our specific test set. However, it is important to note this does not mean that no bias is present in the model. On SpanBERT, we observe that finetuning on the news corpus introduces more preference for straight sentences, while finetuning on the twitter corpus does not introduce this preference. This seems to indicate mainstream media coverage uses language that is more homophobic than queer users on Twitter and that LLMs can ingest and encode this bias when finetuned on news. These results support our hypothesis that the Twitter-tuned models would show less evidence of stereotyping, because queer people talk about themselves in a more nuanced and compassionate way than straight media sources.

\subsection{Coreference Resolution}
In the coreference resolution setting, we evaluate on the OntoNotes \cite{ontonotes} and GAP (Gendered Ambiguous Pronouns, \cite{webster-etal-2018-mind}) datasets. We evaluate the performance of the same 4 LLMs: BERT\_base, BERT\_large, SpanBERT\_base, SpanBERT\_large. We use pre-trained models released by the original model creators, which are usually pre-trained on Wikipedia and other general-domain data. We test three different versions of each model:
\begin{itemize}
    \item Finetune off-the-shelf model on OntoNotes coreference resolution (task only). We call this the ``vanilla'' condition.
    \item Finetune on LGBTQ+ news corpus then OntoNotes (content then task)
    \item Finetune on LGBTQ+ twitter then OntoNotes (content then task)
\end{itemize}

\begin{table*}[!ht]
\begin{centering}
\begin{tabular}{|l|r|r|r|r|}
\hline
 Model & Published & Vanilla & QueerNews & QueerTwitter \\
 \hline
 BERT\_base & 73.9 & \textbf{74.1} & 73.3 & 72.3  \\
 BERT\_large & 76.9 & 76.8 &\textbf{77.0} & 76.3 \\
 SpanBERT\_base & \textbf{77.7} & \textbf{77.7} & 77.5 & 77.2 \\
 SpanBERT\_large & 79.6 & \textbf{80.1} & 79.4 & 79.5 \\
 \hline
 
\end{tabular}
\caption{\label{tab:ontonotes-results} F1 scores on OntoNotes test set for various models and fine-tuning conditions.  Published scores are from \cite{joshi2019coref} and \cite{joshi-etal-2020-spanbert}. }
\end{centering}
\end{table*}
\begin{table*}[!ht]
\begin{centering}
\begin{tabular}{|l|r|r|r|r|r|r|}
\hline
 Model & Vanilla F1 & Vanilla Bias & News F1 & News Bias & Tw. F1 & Tw. Bias \\
 \hline
 \cite{webster-etal-2018-mind} baseline & 70.6 & 0.95 & N/A & N/A & N/A & N/A \\
 BERT\_base & \textbf{82.4} & \textbf{0.97} & 82.3 & \textbf{0.97} & 81.8 & 0.95 \\
 BERT\_large & \textbf{85.6} & 
 \textbf{0.97} & 85.3 & 0.96 & 84.8 & 0.96 \\
 SpanBERT\_base & \textbf{85.5} & \textbf{0.96} & 85.3 & 0.94 & 85.3 & 0.95 \\
 SpanBERT\_large & \textbf{87.7} & \textbf{0.95} & \textbf{87.7} & 0.93 & 87.2 & 0.94 \\
 \hline
 
\end{tabular}
\caption{\label{tab:gap-results} F1 and bias score results. Bias score is the ratio F1 for she/her pronouns to F1 for he/him pronouns. A perfectly unbiased model would have a bias score of 1.}
\end{centering}
\end{table*}

On the OntoNotes test set, our models perform comparably to previously published results. As expected, larger models have generally better performance and less gender bias, and SpanBERT outperforms BERT. We observe relatively little performance degradation from our finetuning. The most severe degradation is observed on BERT\_base finetuned on Twitter, which loses 1.8\% F1 score compared to our vanilla BERT\_base. Loss of performance is even smaller for the larger models and those finetuned on the news corpus. The strong performance on OntoNotes coreference resolution demonstrates that our finetuning does not cause catastrophic forgetting, and that our models are still useful on downstream NLP tasks.

Table~\ref{tab:gap-results} lists results on the GAP dataset for all models. Note that the baseline model in the table is not a BERT-based model, hence the substantially lower scores. The impact of our finetuning on performance is similar to the impact on OntoNotes: very slight performance degradation is observed, with the Twitter corpus having marginally more impact. The interesting result from these experiments was that our finetuning generally did not reduce the gender disparity in model performance. We expected finetuning on queer data to improve gender disparity, but it actually made the gender disparity slightly worse in most cases. For the SpanBERT models, news has a slightly more negative effect on the bias score than Twitter. The reasons for this impact are not yet clear, and further investigation is needed.

\section{Conclusion}
In this work, we present WinoQueer, a new benchmark dataset for quantifying anti-LGBTQ+ bias in large language models. We show that off-the-shelf LLMs show substantial anti-queer bias on our test set. Additionally, we demonstrate that this bias can be mitigated by finetuning the LLMs on a corpus of tweets from the LGBTQ+ community. Finally, we evaluate our models on coreference resolution to show that this finetuning does not cause catastrophic forgetting of the original model weights. There are many promising directions for future work in this area, including developing a more comprehensive version of the WinoQueer benchmark and using it to test a wider variety of LLMs.

\bibliographystyle{acl_natbib}
\bibliography{anthology, ref}

\end{document}